\documentclass[sigconf]{acmart}

\copyrightyear{2025}
\acmYear{2025}
\setcopyright{acmlicensed}
\acmConference[MM '25] {Proceedings of the 33rd ACM International Conference on Multimedia}{October 27--31, 2025}{Dublin, Ireland.}
\acmBooktitle{Proceedings of the 33rd ACM International Conference on Multimedia (MM '25), October 27--31, 2025, Dublin, Ireland}
\acmDOI{10.1145/3746027.3755395}
\acmISBN{979-8-4007-2035-2/2025/10}
\usepackage{svg}
\usepackage{caption}
\usepackage{array}
\usepackage{multirow}
\AtBeginDocument{%
  }

\renewcommand\footnotetextcopyrightpermission[1]{}
\begin{document}

\title{SmartFreeEdit: Mask-Free Spatial-Aware Image Editing with Complex Instruction Understanding}


% \author{Qianqian Sun}
% \email{u3619763@connect.hku.hk}
% %\author{G.K.M. Tobin}
% %\email{webmaster@marysville-ohio.com}
% \begin{center}

\author{Qianqian Sun}
\affiliation{%
  \institution{Institute of Artificial Intelligence(TeleAI), China Telecom}
  \city{Shanghai}
  \country{China}
  \\
  \institution{The University of Hongkong}
  \city{Hongkong}
  \state{}
  \country{China}
}
\email{sqq134050@163.com}
\authornote{Work done during an internship with TeleAI.}
\authornote{Joint first-authors with equal contributions.}

\author{Jixiang Luo}
\affiliation{%
  \institution{Institute of Artificial Intelligence(TeleAI), China Telecom}
  \city{Shanghai}
  \country{China}}
\email{luojx14@chinatelecom.cn}
\authornotemark[2]
\authornote{Project lead}

\author{Dell Zhang}
\affiliation{%
  \institution{Institute of Artificial Intelligence(TeleAI), China Telecom}
  \city{Shanghai}
  \country{China}}
\email{dell.z@ieee.org}
\authornote{Corresponding author.}

\author{Xuelong Li}
\affiliation{%
  \institution{Institute of Artificial Intelligence(TeleAI), China Telecom}
  \city{Shanghai}
  \country{China}}
\email{xuelong_li@ieee.org}
\authornotemark[4]

% \end{center}

% \affiliation{%



\begin{abstract}
% \footnotetext[1]{* means intern at TeleAI and + is corresponding author}

Recent advancements in image editing have utilized large-scale multimodal models to enable intuitive, natural instruction-driven interactions. However, conventional methods still face significant challenges, particularly in spatial reasoning, precise region segmentation, and maintaining semantic consistency, especially in complex scenes. 
% These limitations often stem from the reliance on manual masking, limited instruction-following capabilities, and the fact that existing systems typically focus only on local information within the mask region, leading to suboptimal image generation quality. This local-focused approach fails to preserve structural coherence and semantic integrity across the entire image, especially when dealing with complex or multi-object scenes.
To overcome these challenges, we introduce SmartFreeEdit, a novel end-to-end framework that integrates a multimodal large language model (MLLM) with a hypergraph-enhanced inpainting architecture, enabling precise, mask-free image editing guided exclusively by natural language instructions. The key innovations of SmartFreeEdit include: (1) the introduction of region-aware tokens and a mask embedding paradigm that enhance the model’s spatial understanding of complex scenes; (2) a reasoning segmentation pipeline designed to optimize the generation of editing masks based on natural language instructions; and (3) a hypergraph-augmented inpainting module that ensures the preservation of both structural integrity and semantic coherence during complex edits, overcoming the limitations of local-based image generation.
Extensive experiments on the Reason-Edit benchmark demonstrate that SmartFreeEdit surpasses current state-of-the-art methods across multiple evaluation metrics, including segmentation accuracy, instruction adherence, and visual quality preservation, while addressing the issue of local information focus and improving global consistency in the edited image. 
Our project will be available at 
\url{https://github.com/smileformylove/SmartFreeEdit}.
% \url{ https://anonymous.4open.science/r/SmartFreeEdit-A484/} .

\end{abstract}



% \begin{figure*}[htbp]
% \centering
%   \includegraphics[width=0.8\textwidth]{exp/效果图v5.png}
% \caption{ We propose SmartFreeEdit to address the challenge of reasoning instructions and segmentations  in image editing, thereby enhancing the practicality
% of AI editing. Our method effectively handles some semantic editing operations, including adding, removing, changing objects, background changing and global editing.
% }\label{fig:teaser}
% \Description{}
% \end{figure*}


\maketitle

\begin{figure*}[htbp]
\centering
  \includegraphics[width=0.8\textwidth]{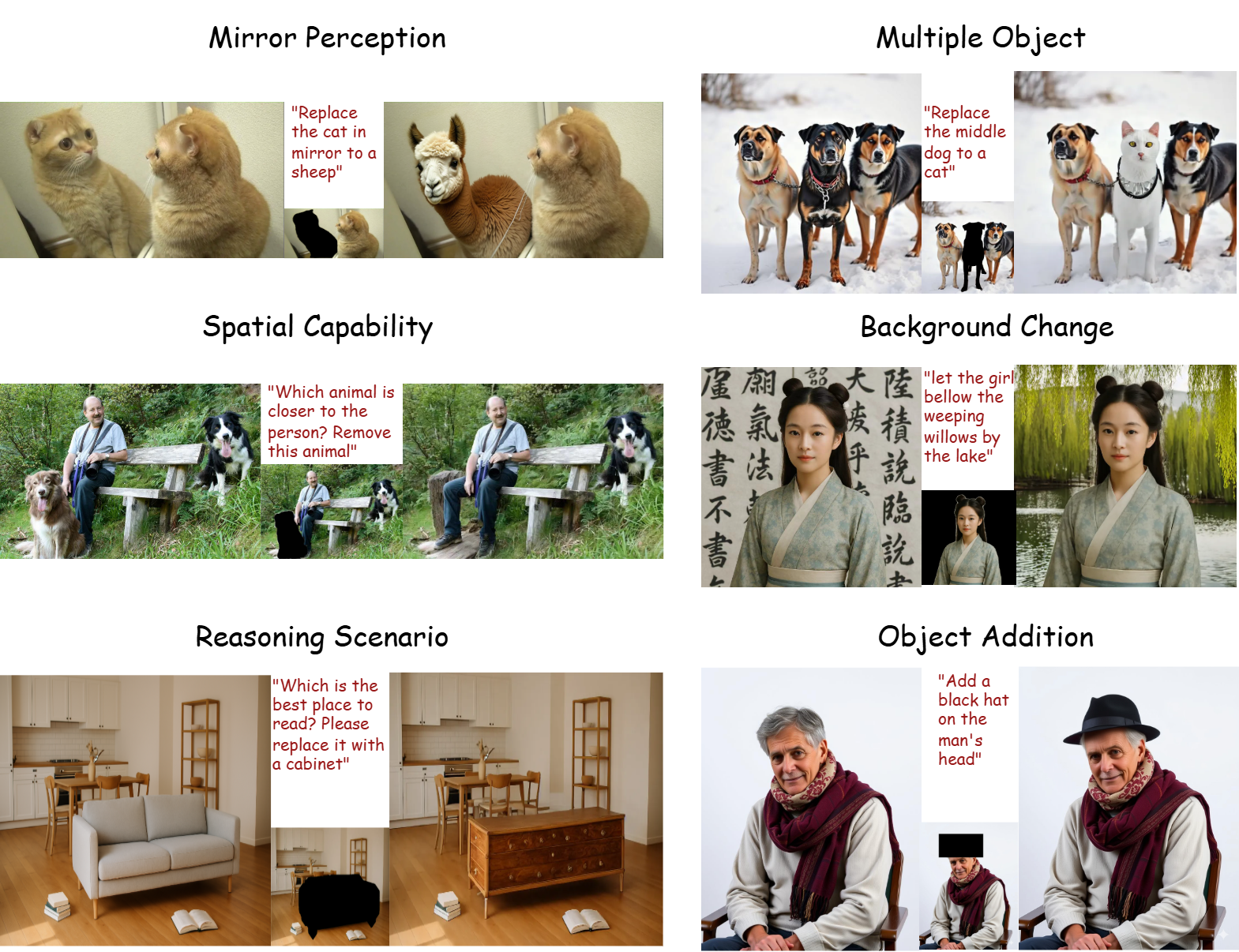}
\caption{ We propose SmartFreeEdit to address the challenge of reasoning instructions and segmentations  in image editing, thereby enhancing the practicality
of AI editing. Our method effectively handles some semantic editing operations, including adding, removing, changing objects, background changing and global editing.
}\label{fig:teaser}
\Description{}
\end{figure*}

\section{Introduction}
With the rapid advancement of visual content processing technologies, image editing has garnered significant attention due to its wide applicability in areas such as content creation, virtual reality, entertainment, and design. In recent years, the emergence of diffusion models~\cite{brooks2023instructpix2pix, epstein2023diffusion, geng2024instructdiffusion, huang2024diffusion, matsunaga2022fine, li2024zone} and autoregressive models \cite{sun2024autoregressive, tian2024visual} has brought about a revolutionary transformation in the fields of image and video generation. These generative models have not only excelled in terms of image fidelity and diversity but have also provided greater control and flexibility for image editing tasks.

% Despite the development of various automated image editing methods, many systems[???] still rely on 
User-provided hand-drawn masks are used to indicate the editing area, while this approach offers a certain level of interactivity and flexibility, it places high demands on the operational skills and time investment, especially in complex or large-scale scenarios, limiting its practical applicability. While systems like MagicQuill~\cite{liu2024magicquill} have improved the editing experience by integrating modules such as editing processors, drawing aids, and idea collectors, they still rely heavily on precise user input, which presents challenges in efficiency and scalability.

Instruction-based image editing, driven by natural language commands, has opened a new direction for the intelligent development of image editing. InstructPix2Pix~\cite{brooks2023instructpix2pix} was the first to achieve image editing through natural language, marking a shift from traditional interactive tools to intelligent assistants. Subsequently, a series of studies introduced large language models (LLMs) to enhance the ability of understanding and execution of complex instructions, advancing the field of multimodal editing. Recently, Gemini 2.0 Flash~\cite{deepmind2024gemini}, Google latest multimodal large model, has demonstrated exceptional cross-modal reasoning capabilities. SmartEdit~\cite{huang2024smartedit} systematically analyzed these issues and pointed out that existing models are limited by two key factors: (1) Text encoders like CLIP~\cite{radford2021learning} struggle to accurately extract editing intent from complex language, and (2) Training data with synthetic images leads to challenges with complex scenes and high dependency on image-text alignment. Furthermore, many models integrate MLLMs and segmentation mechanisms like Grounded-SAM~\cite{ren2024grounded}, which intergres Grounding-DINO~\cite{liu2024grounding} and SAM~\cite{kirillov2023segment}. However, Grounded-SAM faces difficulties in handling reasoning tasks that rely on implicit information such as spatial, functional, and contextual relationships. However, despite performing well in simple tasks, existing systems still face significant limitations in handling complex scenarios such as: 1) \textbf{Complex instruction understanding}: in multi-object scenes, the model must recognize target objects with similar appearances and distinguish their attributes (e.g., color, material, spatial location). For instance, "Make the middle animal in the image into a cute cat" requires editing model to recognize the exactly position of all animals. 2) \textbf{Reasoning with world knowledge}: inferring the position of a sofa in a room based on a prompt like "what is the best place to read in this layout," which requires contextual, functional and spatial reasoning.
% \begin{itemize}
% \setlength{\itemindent}{0pt}  % Remove extra indentation
% \setlength{\leftmargini}{0pt} % Remove indentation from list items
% \item \noindent\textbf{Complex instruction understanding.} In multi-object scenes, the model must recognize target objects with similar appearances and distinguish their  attributes (e.g., color, material, spatial location). For instance, "Make the middle animal in the image into a cute cat" requires editing model to recognize the exactly position of all animals.
% \item  \noindent\textbf{Reasoning with world knowledge.} For example, inferring the position of a sofa in a room based on a prompt like "what is the best place to read in this layout," which requires contextual, functional, and spatial reasoning.
% \end{itemize}

Thus segmentation-based image editing is proposed as a core technology. Unlike traditional semantic segmentation, reasoning segmentation focuses on understanding "implicit semantic instructions," i.e., identifying the required editing regions through concise but semantically rich language expressions such as functional descriptions, spatial relationships, and contextual reasoning. This method combines visual perception with knowledge reasoning to automatically generate high-quality masks, significantly improving target localization and task execution in complex editing scenarios.
Moreover, when generating and repairing image masks, we found that traditional image inpainting models, such as LaMa~\cite{suvorov2022resolution} and Paint-by-Example~\cite{yang2023paint}, focus mainly on filling in local texture information within the masked region. These methods generally perform well for simple or background-consistent images, producing visually continuous local content. However, they often lack the ability to model the overall structure and semantic relationships within the image. In scenes with multiple objects, complex layouts, or long-range dependencies, such models can introduce edge distortions, semantic inconsistencies, and structural artifacts. 

To overcome above all issues, we propose a structure-aware inpainting mechanism in \textbf{SmartFreeEdit}, which introduces hypergraph reasoning to model the high-order relationships between key objects, regions, and their spatial-semantic interactions. Unlike traditional graph structures that describe point-to-point relationships, hypergraphs enable the modeling of one-to-many or many-to-many semantic and spatial connections, ensuring global consistency during the editing process and maintaining structural integrity in the final repaired image. In summary, SmartFreeEdit is an end-to-end system for instruction-based image editing, specifically designed for complex reasoning scenarios. Our system consists of three key modules:
\begin{itemize}
    \item \textbf{MLLM-driven Promptist.} An agent that decomposes user input into editing objects, categories, and target prompts, enabling structured understanding of task intent.
    \item \textbf{Reasoning Segmentation.} By transforming editing objects into image query expressions and combining multimodal knowledge reasoning, this module generates semantically consistent masks to aid subsequent image operations.
    \item \textbf{Inpainting with Hypergraphs.} A structure-aware image inpainting model that uses hypergraph reasoning to enhance global consistency, maintaining the image's overall structure and ensuring high-quality repairs.
\end{itemize}
\section{Related Works}
\subsection{Instruction-based Image Editing}
The InstructPix2Pix~\cite{brooks2023instructpix2pix} framework represents a significant advancement in the field of image editing. It introduces a guided image editing model trained on a large synthetic dataset of instruction-based edits. 
% Instruction-driven image editing methods have garnered increasing attention in the research community due to their alignment with human image editing practices, where users typically provide high-level instructions for image manipulation. 
MagicBrush~\cite{zhang2023magicbrush} enhances this approach by fine-tuning InstructPix2Pix on a manually annotated dataset using an online editing tool. %In addition, %InstructDiffusion~\cite{geng2024instructdiffusion} further extends this framework by building and integrating traditional computer vision tasks, enhancing the versatility in performing diverse image edits.
% In recent years, MLLMs have emerged as powerful tools in the image editing domain. These models enable tasks such as text-to-image generation, prompt refinement, and image quality assessment. One notable approach, 
MGIE~\cite{fu2023guiding} utilizes MLLMs to generate more expressive and concise instructions, thereby enhancing instruction-based image editing workflows. SmartEdit~\cite{huang2024smartedit} introduces a bidirectional interaction module that facilitates mutual understanding between text representations output by Vision-Language Models (VLMs) and the input image features, ensuring better coherence in the editing process.
Furthermore, FlexEdit~\cite{nguyen2024flexedit} integrates MLLMs to comprehend image content, masks, and textual instructions, enhancing the ability to perform complex edits based on nuanced input. GenArtist~\cite{wang2024genartist} leverages MLLM capabilities to break down complex tasks, guide tool selection, and perform step-by-step validation, enabling systematic image generation, editing, and self-correction. FireEdit~\cite{zhou2025fireedit}, on the other hand, employs region-aware VLMs to interpret editing instructions in complex scenes, providing more context-aware modifications. Lastly, BrushEdit~\cite{li2024brushedit} incorporates MLLMs, detection models, and BrushNet~\cite{ju2024brushnet} inpainting models into a training-free, agent-cooperative framework, allowing users to input simple natural language instructions while ensuring seamless image editing. 

Our model is also inspired by the application of the MLLMs mentioned above in the field of image editing. By combining the powerful analysis and reasoning capabilities of MLLM, our system is able to more precisely identify and understand the type of editing, the object of editing, and its context, thereby optimizing the effectiveness of image editing.

\subsection{Reasoning Segmentation}
% The inference segmentation task aims to accurately segment specific objects mentioned in the query text through complex instructions. In recent years, models such as 
Segment Anything Model(SAM)\cite{kirillov2023segment} have gained widespread attention in the community due to their precise segmentation capabilities and strong zero-shot functionality. However, the performance of SAM in natural language prompts still requires improvement, particularly in handling segmentation tasks with complex semantics. To address this issue, Grounded-SAM\cite{ren2024grounded} integrates open-set detector models, such as Grounding-DINO\cite{liu2024grounding}, with promptable segmentation models like SAM, effectively overcoming the challenge of open-set segmentation. However, its spatial awareness and ability to segment complex scenes remain limited.
With advancements in technology, the grounding ability of MLLM has been continuously enhanced, enabling users to directly point to key objects or regions in images. Specifically, Kosmos-2 \cite{peng2023kosmos} and VisionLLM\cite{wang2023visionllm} train on image-text pairs to enable dialogue-based understanding of specific regions. Additionally, GPT4RoI\cite{zhang2025gpt4roiinstructiontuninglarge} uses spatial bounding boxes and RoIs to align regions with textual prompts, further improving regional accuracy in image segmentation. LISA \cite{lai2024lisa} proposed an inference segmentation method based on visual-language model embeddings and SAM decoders, allowing the model to perform precise segmentation in complex scenes using effective attention mechanisms. Meanwhile, GlaMM\cite{rasheed2024glamm} successfully advances reasoning mask segmentation and dialogue models through the collaborative work of its global image encoder, region encoder, LLM module, aligned image encoder, and pixel decoder, achieving scene-level understanding, region-level understanding, and pixel-level alignment.

Building upon these developments, our approach combines the strengths of LLM and SAM, aiming to activate the reasoning and spatial awareness capabilities with a small inference training set. By integrating inference segmentation with image inpainting, our model is capable of performing highly precise image editing tasks with reasoning capabilities.

\begin{figure*}[htbp]
    \centering
    \includegraphics[width=0.8\linewidth]{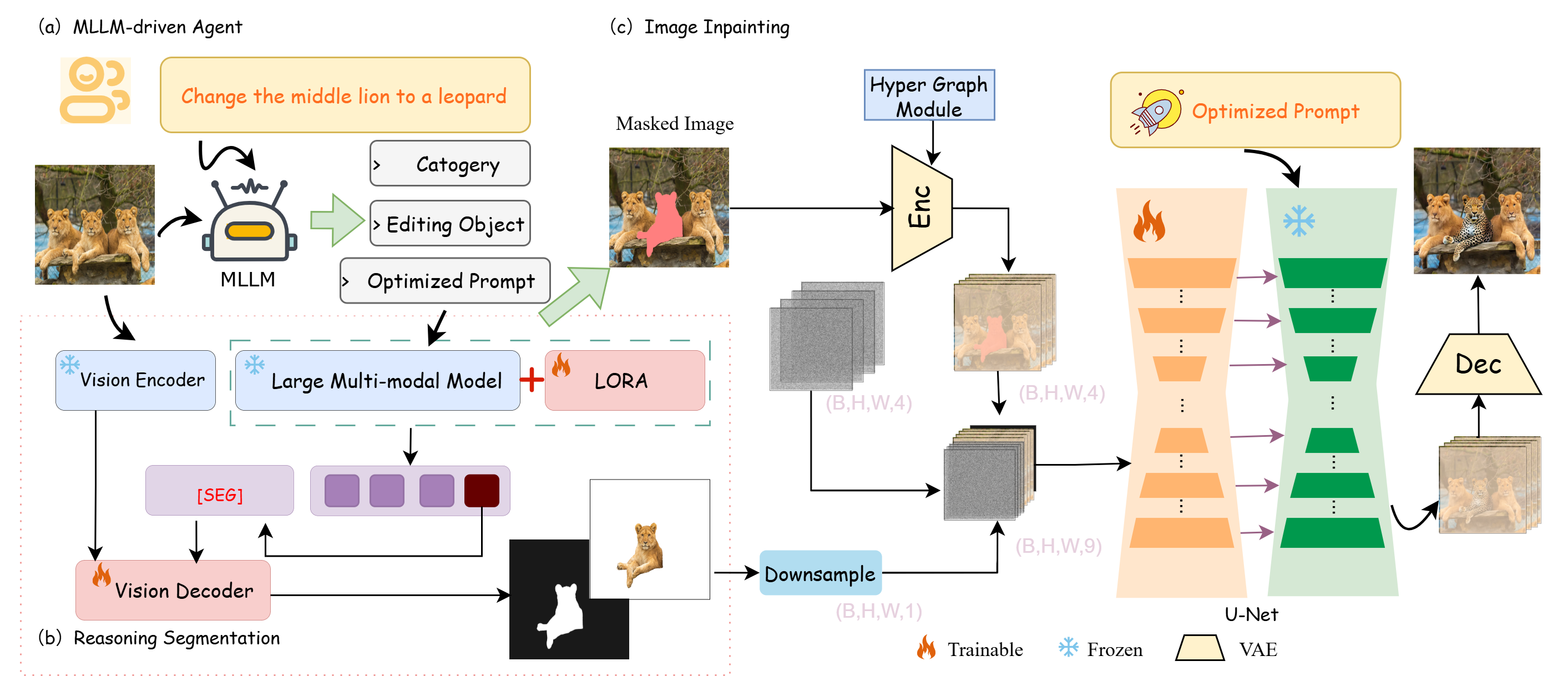}
    \caption{Architecture Overview of SmartFreeEdit for Reasoning Complex Scenarios Instruction-Based Editing. Our SmartFreeEdit consists of three key components: 1) An MLLM-driven Promptist that decomposes instructions into Editing Objects, Category, and Target Prompt. 2) Reasoning Segmentation converts the prompt into an inference query and generates reasoning masks. 3) An Inpainting-based Image Editor using the hypergraph computation module to enhance global image structure understanding for more accurate edits.}
    \label{fig:pipeline}
\Description{}
\end{figure*}
\subsection{Image Inpainting}
% The primary objective of image inpainting models is to restore or fill in occluded, missing, or damaged parts of an image, ensuring that the repaired image appears natural and coherent.
Traditional image inpainting methods, such as Autoencoders~\cite{mou2024t2i}, Autoregressive Transformers~\cite{wan2021high}, and Generative Adversarial Networks (GANs) \cite{sargsyan2023mi,xu2023image}, typically rely on manually designed features for the inpainting process. However, these methods exhibit limited performance, especially in the restoration of details in complex scenes. 
% The main reason for this limitation lies in their inability to fully exploit contextual information, particularly the relationships between mask boundaries and unmasked regions.
% With the advent of Diffusion Models, image inpainting techniques have seen significant improvement. 
Early inpainting methods~\cite{avrahami2023blended} often employed sampling strategies to replace unmasked regions, performing simple inpainting tasks such as region filling. However, these methods fail to adequately perceive the mask boundaries and the surrounding contextual information, resulting in suboptimal inpainting results. To address this issue, some studies have introduced fine-tuning techniques to optimize the inpainting process. For example, Stable Diffusion Inpainting (SDI)~\cite{rombach2022high} adjusts the input of the U-Net architecture to include both the mask, the masked image, and the noise latent variables, thereby enabling the inpainting process to better capture mask boundary information, ultimately enhancing the inpainting results.
Moreover, SmartBrush~\cite{xie2023smartbrush} combines text and shape guidance and enhances the perceptual ability of the diffusion U-Net through object mask prediction. ControlNet~\cite{zhang2023adding}introduces additional control information, leveraging more prior knowledge in the inpainting task to improve the controllability of the results, although it is not suitable for pixel-level inpainting feature injection. BrushNet~\cite{ju2024brushnet}employs a dual-branch strategy for mask-guided insertion and trains the model using global prompts, allowing efficient inpainting without the need for additional features designed manually. 

To overcome the existing limitations of global perception and spatial relationship modeling in current inpainting methods, we propose an image inpainting model based on hypergraph computation. This module facilitates information sharing between distant pixels or regions, enhancing the overall understanding of features and further improving the image-inpainting performance.

\section{Methodology}
The SmartFreeEdit framework integrates multiple technological modules to achieve instruction editing with reasoning and spatial awareness capabilities. In Section \ref{mllm}, we provide a detailed explanation of how the system relies on an MLLM (like GPT4o~\cite{openai2024gpt4o}) for instruction parsing and text prompt optimization. Section \ref{rs} describes the specific process of generating the reasoning segmentation mask. In Section \ref{hyper}, we delve into the structure and implementation details of the image repair model based on the hypergraph computation module. Our system efficiently and accurately completes tasks ranging from simple image modifications to intricate scene detail adjustments.
\subsection{MLLM-driven Promptist}
\label{mllm}
The core of the SmartFreeEdit framework is the MLLM-driven Promptist, which plays a crucial role in analyzing, extracting, and optimizing user instructions for complex image editing tasks. This agent is responsible for processing natural language input and transforming it into actionable tasks through the following steps:

\noindent\textbf{Instruction Analysis.} First, the MLLM parses the user natural language input, handling tasks ranging from simple object editing to complex scene adjustments. Leveraging the multimodal capabilities of GPT-4o, the agent understands the context of the instructions and extracts relevant information such as the editing object (e.g., the object to be replaced or added) and the type of edit (e.g., 'Remove', 'Addition', 'Replace', 'Background', 'Global'). In the case of "Addition" tasks, the agent determines the specific size and position of the inserted part based on the instructions and the given image size.

\noindent\textbf{Prompt Optimization.} After analyzing the instructions and identifying the editing components, the agent further optimizes the prompt information to ensure accurate task execution. Addressing the issue in the dual-branch inpainting model BrushNet \cite{ju2024brushnet}, where the removal of the cross-attention layer causes a lack of hierarchical control and semantic understanding, leading to a mismatch between the generated image and the text prompt, the agent improves prompt optimization. It generates more precise and refined local text prompts based on the type of edit. For "Replace" or "Addition" tasks, the prompts refine the description of the editing object to accurately reflect the required changes; for "Remove", "Background", or "Global" editing tasks, the prompts include specific global information about what should be removed or altered. By refining details, adding contextual information, and ensuring clear task definitions for downstream processing, the MLLM enhances the quality and effectiveness of the prompts.

Through these steps, the SmartFreeEdit framework significantly improves the accuracy of image editing task execution and the understanding of user instructions, enabling efficient and precise image processing.

\subsection{Reasoning Segmentation}
\label{rs}
As shown in the Figure \ref{fig:pipeline}, our architecture is similar to LISA, integrating two foundational models to achieve high-precision segmentation output: 1) MLLM as the aligned visual-language cognitive module, and 2) SAM as the segmentation base model, which segments the target object from the input image based on user instructions. To embed reasoning segmentation into image editing, we also introduce the segmentation token \(<seg>\) into the main vocabulary, which signifies a request for segmentation output.

First, the user instructions are converted into a structured query consistent with the reasoning segmentation format. This query is capable of handling not only simple phrases (for example, '\textit{an apple}'), but also complex and implicit descriptions. For example, through MLLM-driven Promptist analysis, the model understands that the object to be segmented is '\textit{food contains the most vitamin}' and transforms this into the query: '\textit{Please segment the food that contains the most vitamin in the image}'. This process enables the query to handle more complex semantic structures.

When the model receives this query along with the corresponding image, it generates a text response, which may include the <seg> token. This token acts as a placeholder for the target object to be segmented. The hidden state extracts the <seg> token and processes it through a multilayer perceptron (MLP), converting it into an embedding that captures the essential features of the object to be segmented. Meanwhile, the visual encoder processes the image and extracts visual features. This approach allows the model to perform fine-grained reasoning based on both the textual input and the visual cue, enabling it to understand the nuances of the request.

The semantic embedding from the <seg> token and the visual features from the image are combined as our information sources. These two sources are input into the segmentation decoder, which generates a binary segmentation mask for the requested object. This end-to-end process enables the model to generate accurate object masks directly from text descriptions, without requiring additional human intervention or predefined object categories.

The segmentation loss is denoted as \(L_{mask}\), integrating the binary cross-entropy (BCE) loss per pixel alongside the Dice coefficient loss (DICE). The segmentation loss function as follows:
\begin{equation}
    L_{mask} = \lambda _{bce}BCE(\hat{M},M )+\lambda _{dice}DICE(\hat{M},M) 
\end{equation}
where \(\lambda_{bce}\) and \(\lambda_{dice}\) are hyperparameters controlling the relative importance of each term in the overall loss. \(\hat{M}\) and \(M\) stand for the predicted mask and the ground-truth of token \(<seg>\) respectively. Every projected embedding of \(<seg>\) token is fed into the decoder successively to generate the corresponding  \(\hat{M}\). And the text generation Loss \(L_{txt}\) computed using cross-entropy (CE) between the predicted text output \(\hat{y_{txt}}\)  and the ground-truth text target \(y_{txt}\) ensures that the model generates text that aligns with the input query and follows the required text-based reasoning. The loss is fomulated as:
\begin{equation}
    L_{txt} = CE(\hat{y_{txt}}, y_{txt})
\end{equation}

Our SmartFreeEdit is trained by combining the tasks and data with task learning. Thus, the overall loss \(L\) consists of \(L_{txt}\), \(L_{mask}\):
\begin{equation}
    L = \lambda_{txt}L_{txt} + \lambda_{mask}L_{mask}
\end{equation}
This formulation encourages the model to simultaneously optimize its text generation capabilities and segmentation accuracy, ensuring that the model learns to respond to implicit user requests with both textual descriptions and fine-grained visual outputs.

\subsection{Inpainting with Hypergraphs}
\label{hyper}

\begin{figure}[htbp]
    \centering
    \includegraphics[width=\linewidth]{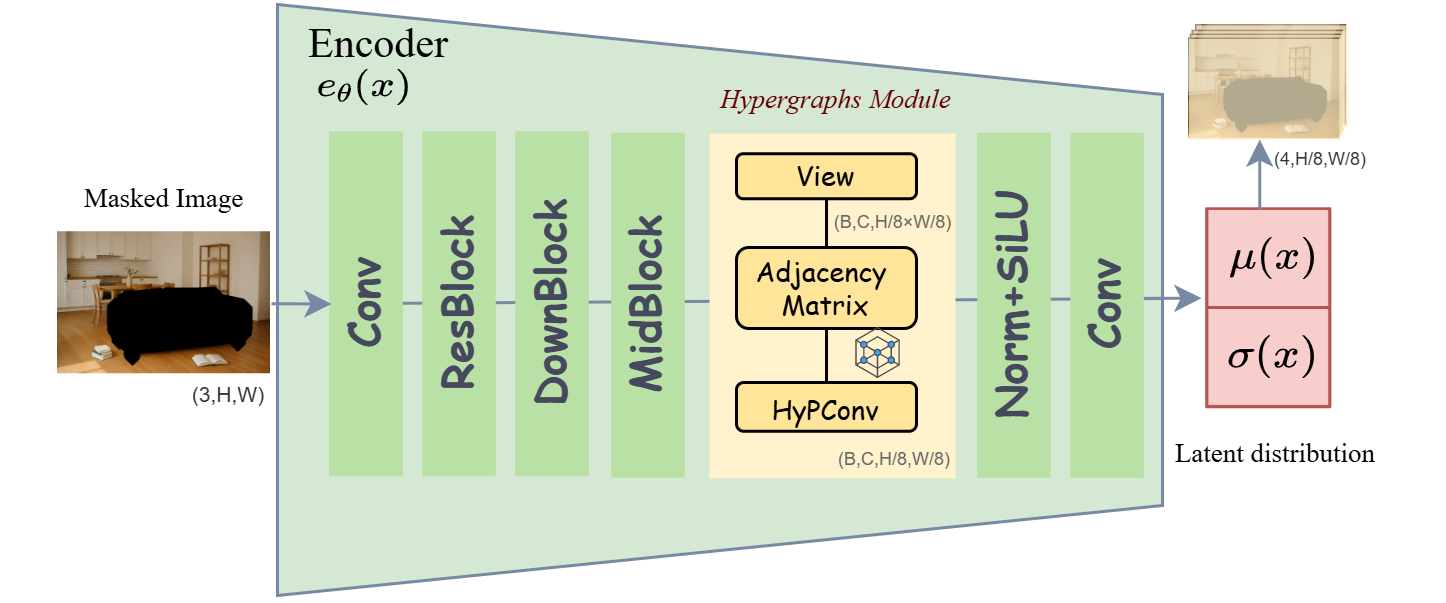}
    \caption{The Proposed Architecture of Hypergraph Module in Encoder for Image Inpainting. The masked image is processed through convolutional layers, residual blocks, and a downsampling block, followed by a hypergraph module that aggregates contextual information through hypergraph convolution (HyPConv). The resulting latent distribution is used for image restoration.}
    \label{fig:hyper}
    \Description{}
\end{figure}
Hypergraph-based inpainting uses hypergraph theory to reconstruct missing or occluded parts of an image by utilizing the spatial and contextual information of pixels or regions. Unlike traditional graph-based methods, where nodes represent pixels and edges represent relationships between them, hypergraphs allow for more flexible connections by capturing relationships among multiple nodes within a single hyperedge. This approach is better suited for modeling complex interactions in high-dimensional spaces, such as images, where multiple neighboring pixels need to be considered together for effective inpainting.

In image inpainting tasks, the goal is to fill in the missing or occluded regions of an image based on the available contextual information. To this end, we introduce the concept of graph neural networks (GNNs) into the intermediate feature space of a Variational Autoencoder (VAE), constructing a Euclidean distance-based graph where each pixel in the image is represented as a vertex \(V = \left \{v_{1}, v{2}, ..., v_{n}\right \}\) in the hypergraph and a set of hyperedges \(E=\left \{e_{1}, e_{2},..., e_{n}\right \}\) connects a group of vertices (pixels). By performing matrix multiplication with the adjacency matrix and features, we implement information propagation between nodes and edges using a \(v2e\longrightarrow e2v\) mechanism, which involves two rounds of feature exchange in the graph structure.

In the hypergraph module, we calculate the Euclidean distance between feature points, and if the distance is smaller than a set threshold, an edge is formed, creating a "local perception graph." Hypergraph convolution (HyPConv) is then applied to propagate information within the hypergraph. This operation extends traditional graph convolution, not only aggregating information from neighboring nodes connected by edges but also considering all vertices connected by hyperedges.

Specifically, the standard graph convolution operation on a node  \(v_{i}\)is given by:
\begin{equation}
   h_{i}^{(k+1)} = \sigma (\sum_{j\in N(i)}^{} w_{ij}h_{j}^{(k)}+b_{i} )
\end{equation}
where \(h_{i}^{(k)}\) is the feature of node \(i\) at the 
\(k\)-th iteration, \(N(i)\) is the set of neighbors of node \(i\), and \(w_{ij}\) is the weight between nodes \(i\) and nodes \(j\). 

For the hypergraph convolution, the operation is generalized to:
\begin{equation}
    h_{i}^{(k+1)} = \sigma (\sum_{e\in \varepsilon(i)}^{} w_{e}h_{e}^{(k)}+b_{i} )
\end{equation}
where \(h_{e}^{(k)}\) is the feature of the hyperedge \(e\) at the \(k\)-th iteration, and \(\varepsilon(i)\) is the set of hyperedges that contain node \(i\). This allows each pixel (node) to aggregate information from a set of connected pixels (nodes) through hyperedges.

As shown in Figure \ref{fig:hyper}, we insert the hypergraph module after the middle block of the VAE encoder. The features are first flattened into the format \(\left [ B,C,HW \right ] \), the graph structure is constructed and processed, and then the features are restored to their original shape \(\left [ B,C,H,W \right ] \). This can be seen as an enhancement of the intermediate features. Similarly, in the VAE decoder, the hypergraph module is also inserted after the middle block for processing. In this way, the hypergraph module effectively enhances the feature recovery capability within the VAE framework, optimizing the image inpainting performance.
\begin{figure*}[htbp]
    \centering
    \includegraphics[width=0.85\linewidth]{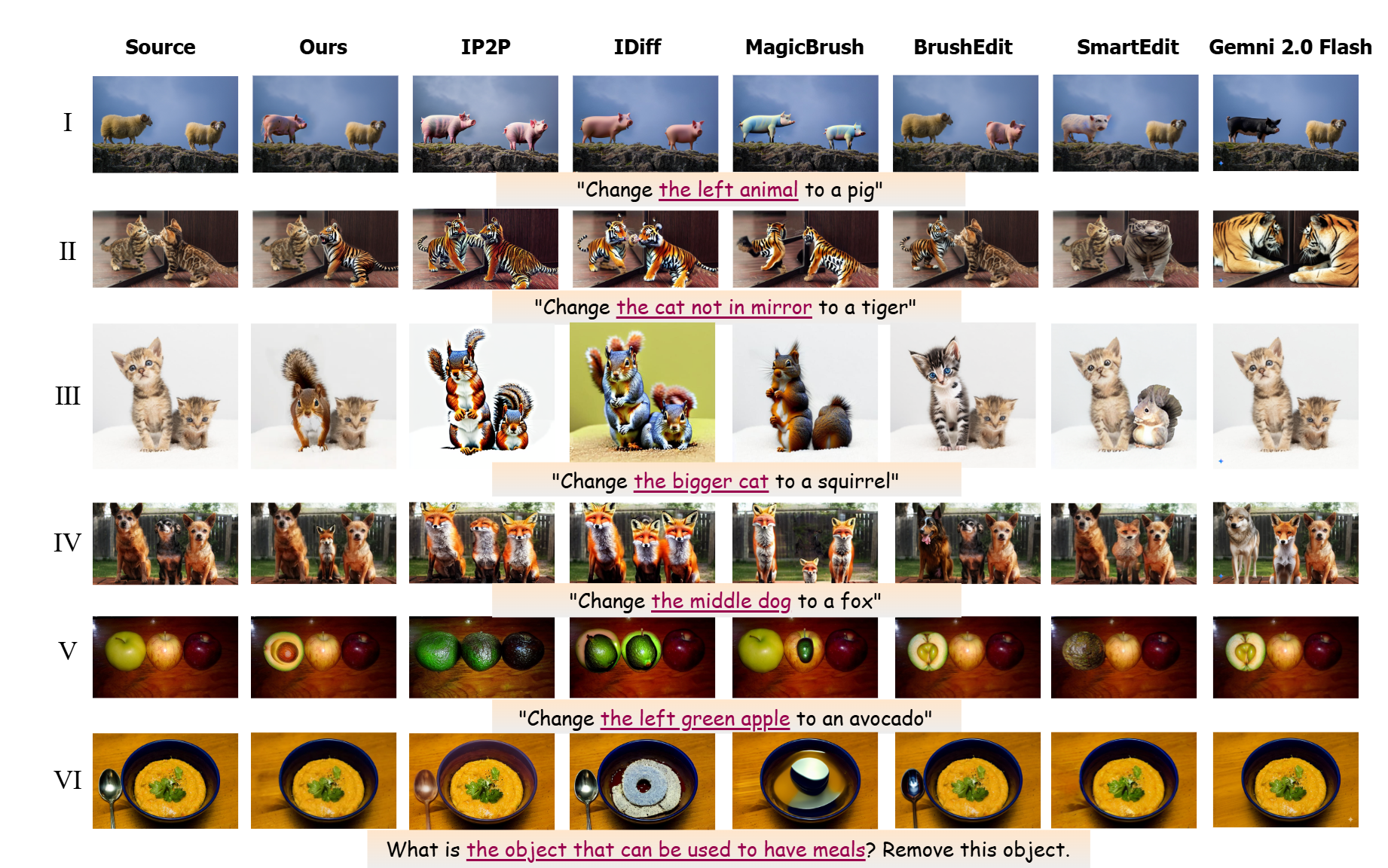}
    \caption{Qualitative comparisons of SmartFreeEdit on Reason-Edit with previous instruction-based image editing methods including InstructPix2Pix (IP2P), InstructDiffusion(IDiff), MagicBrush, BrushEdit, SmartEdit(13B) and latest Gemini 2.0 Flash. Mask-free methods don't require additional mask input, where we take these methods with same instructions as baselines for comparison and our approach demonstrates superior editing capabilities in complex scenarios. }
    \label{fig:Quality-reason}
    \Description{}
\end{figure*}
\section{Experiments}
\begin{table*}[htbp]
  \caption{Quantitative comparison of SmartFreeEdit with existing methods on Reason-Edit.}
  % Red denotes the best.Underline denotes the second best}
  \label{tab:comparision}
  \begin{tabular}{c|ccccc|ccccc}%l
    \toprule
    % \multirow{2}{*}{Methods} & \multicolumn{5}{c}{Understanding Scenarious}  & \multicolumn{5}{c}{Reasoning Scenarios}\\
\multirow{2}{*}{Methods} & 
\multicolumn{5}{c}{Understanding Scenarios} & \multicolumn{5}{c}{Reasoning Scenarios} \\
%\cmidrule(lr){2--11}
    \cmidrule(lr){2-11} 
    &PSNR↑&\(LPIPS_{\times 10^{3}}\downarrow\)&SSIM↑&CLIPSim↑&Ins-align↑&PSNR↑&\(LPIPS_{\times 10^{3}}\downarrow\)&SSIM↑&CLIPSim↑&Ins-align↑\\    
    \midrule
    InstructPix2Pix\cite{brooks2023instructpix2pix} & 21.57&89&0.72&22.76&0.54
    &21.01	&14.00&	0.64&	20.02&0.167
\\
    InstructDiffusion\cite{geng2024instructdiffusion}& 13.82 &277.30&0.59&\textbf{\color{red}{25.03}}&0.51 & 
    19.13&	16.50&	0.63&	20.23&0.42
\\
    MagicBrush\cite{zhang2023magicbrush} & 16.85 & 185.20&0.66 & 23.44
& 0.56&19.85&	16.21&	0.65&	19.62&	0.33\\
    BrushEdit\cite{li2024brushedit}&28.45 & 61.90& 0.90&20.19&0.38
    &30.45&\textbf{\color{red}{ 8.01}} &0.88&18.47&0.18
\\
    SmartEdit-7B\cite{huang2024smartedit}&22.05 &87.00&0.73&23.61&	0.71&
    25.26&	55.67	&0.74&	20.95	&0.79
    \\
    SmartEdit-13B\cite{huang2024smartedit}&23.60&68.01&0.75	&23.54	&0.77
    &25.76&	51.43&	0.74&20.78&\textbf{\color{red}{0.82}}\\
    Gemini 2.0 Flash	\cite{deepmind2024gemini}&20.06	&138.76&	0.73	&22.98&	0.57&22.47&105.02&0.73&20.80&0.55\\
    Ours &\textbf{\color{red}{28.99}}&\textbf{\color{red}{51.49}}&\textbf{\color{red}{0.92}}&24.43&\textbf{\color{red}{0.86}}&
    \textbf{\color{red}{31.27}}& 	47.00&	\textbf{\color{red}{0.92}}&\textbf{\color{red}{21.46}}&	0.70\\
  \bottomrule
\end{tabular}
\end{table*}

\subsection{Implementation Details}

\begin{table*}[htbp]
  \caption{Quantitative comparison on BrushBench(* with blending operation)}
  \label{tab:BrushBench}
  \begin{tabular}{c|ccc|ccc|c}%l
    \toprule
   \multirow{2}{*}{Models} & \multicolumn{3}{c}{ Image Quality} &  \multicolumn{3}{c}{ Masked Region Preservation}& Text Align\\ 
    % \hline
    \cmidrule(lr){2-8} 
   &\(IR_{\times 10}\uparrow \)&\(HPS_{\times 10^{2}}\uparrow \)&\(AS \uparrow\)&
   \(PSNR\uparrow\)&\(LPIPS_{\times 10^{3}}\downarrow\)&\(MSE_{ \times 10^{3
}}\downarrow\)&\(CLIPSim\uparrow\)\\
    \midrule
    BLD\cite{avrahami2023blended}& 5.45& 23.42 &5.43 &24.89& 98.79& 4.46& 26.31\\
    SDI\cite{rombach2022high}&8.91&25.70&5.96 &30.61&20.88&1.08  &25.89\\
     HDP\cite{manukyan2023hd}&  10.14&26.26 &6.19 &22.07&25.06&39.69&25.92\\
     CNI\cite{zhang2023adding}& 9.33&25.65&6.12&21.82&62.89&44.00&25.73\\
 CNI*\cite{zhang2023adding}&11.94 &27.62& 6.26& 31.38&19.34&4.85&25.95\\

 BrushNetX*\cite{li2024brushedit}&12.76	&28.31	&6.27	&\textbf{\color{red}{31.76}}&19.33	&\textbf{\color{red}{0.82}}&	26.57\\
 Ours* &\textbf{\color{red}{13.39}}	&\textbf{\color{red}{28.50}}&	\textbf{\color{red}{6.48}}&	31.67&\textbf{\color{red}{19.13}}&	0.83	&\textbf{\color{red}{26.80}}
\\

  \bottomrule
\end{tabular}

% \noindent \multicolumn{8}{l}{\textit{* with blending operation}}

\end{table*}

All comparison experiments for image inpainting were conducted using the NVIDIA A800 GPU. The base models used were Stable Diffusion v1.5~\cite{runway2022sdv15} and Stable Diffusion XL~\cite{runway2022sdvXL}, with 50 steps and a guidance scale of 7.5. To ensure fair comparisons, we performed inference using the recommended hyperparameters for each method. Additionally, our hypergraph-based inpainting model and ablation experiments were trained for 100,000 steps on 8 NVIDIA A800 GPUs, with a learning rate set to 1e-5 and Our training data consists of 25\(\%\) of the BrushData~\cite{ju2024brushnet}.

\subsection{Benchmark and Metrics}
To comprehensively evaluate the performance of SmartFreeEdit, we conducted experiments on image editing and image inpainting benchmarks separately:

\noindent \textbf{Image Editing.} We evaluate our performance in complex understanding and reasoning scenarios using the Reason-Edit benchmark from SmartEdit~\cite{huang2024smartedit}, comparing our method against all baselines. Reason-Edit consists of 219 image-text pairs, distributed across seven reasoning and understanding scenarios: left-right, relative size, mirroring, color, multiple objects, reasoning, and addition. Each image is accompanied by three annotations: the source image, the editing instruction, and the editing mask.

\noindent \textbf{Image Inpainting.} The newly launched BrushBench \cite{ju2024brushnet}, designed to evaluate diffusion-based image generation models, contains 600 images, each with a human-annotated mask and captions. The images cover a balance of natural and artificial scenes, including paintings, and ensure an even distribution of different categories such as human, animals, indoor and outdoor environments. This classification makes the evaluation more fair and comprehensive, and helps to measure model performance objectively and consistently.

\noindent \textbf{Metrics.} In the image editing task, we used six evaluation metrics, focusing on the retention of masked areas and the alignment of text with the edited content. Specifically, standard peak signal-to-noise ratio (PSNR), learned perceptual image patch similarity (LPIPS), structural similarity index (SSIM), and mean square error (MSE) are used to analyze the performance of the masked region between the generated image and the original image. In addition, we have introduced a CliP-based similarity indicator to assess the semantic consistency between the edited image and the target modification. In particular, we use "whether the edited image complies with the directive" as a core evaluation criterion. In order to further improve the evaluation process, we combined the manual evaluation as a reference index, collected and averaged the evaluation results of 8 different taggers, and finally obtained the 'Ins-align' evaluation index to measure the consistency of image editing and user instructions.

When evaluating image inpainting models, we also introduce complementary metrics such as image Reward (IR), HPS v2 (HPS), and aesthetic score (AS), which are designed to be as consistent as possible with human perceptual judgments.
\subsection{Quantitative and Qualitative Comparison of Image Editing}
In qualitative experiments on Figure \ref{fig:Quality-reason}, SmartFreeEdit performed well on image understanding and inference-driven tasks, consistently producing high-quality image edits consistent with the semantics of the instructions. Compared to previous models such as InstructPix2Pix\cite{brooks2023instructpix2pix}, MagicBrush\cite{zhang2023magicbrush}, and InstructDiffusion\cite{geng2024instructdiffusion}, their simpler encoders struggle to accurately capture the semantics of instructions in complex scenarios, especially in tasks requiring deep reasoning. Our model gives full play to the reasoning ability of MLLM, and effectively identifies and edits target objects. 

In spatial inference tasks, SmartFreeEdit was able to accurately handle complex scenes, while other models, such as BrushEdit and Gemini 2.0 Flash, made errors in object recognition and editing. For example, in the cat mirror replacement task in Figure II, they failed to understand the command; In Figure III, faced with multiple targets of the same type, they cannot locate and only edit the animals in the middle. Compared to SmartEdit, our approach not only maintains high semantic consistency, but also significantly improves image quality.

In the quantitative experiments in Table \ref{tab:comparision}, we observed that InstructDiffusion achieved the highest CLIPSim in understanding tasks, but at the cost of poor image background retention. In comparison to previous models, our method achieved a higher editing success rate, ensuring both content preservation and alignment with the editing instructions. In inference tasks, while BrushEdit showed similar image quality to ours, it lacked sufficient reasoning capability. Its output contained numerous areas identical to the original image, giving the illusion of high quality. Quantitatively, our model demonstrated superior image editing performance, outperforming both SmartEdit and Gemini 2.0 Flash in terms of overall image quality.
\subsection{Quantitative and Qualitative Comparison of Image Inpainting}

\begin{figure*}[htbp]
    \centering
    \includegraphics[width=0.86\linewidth]{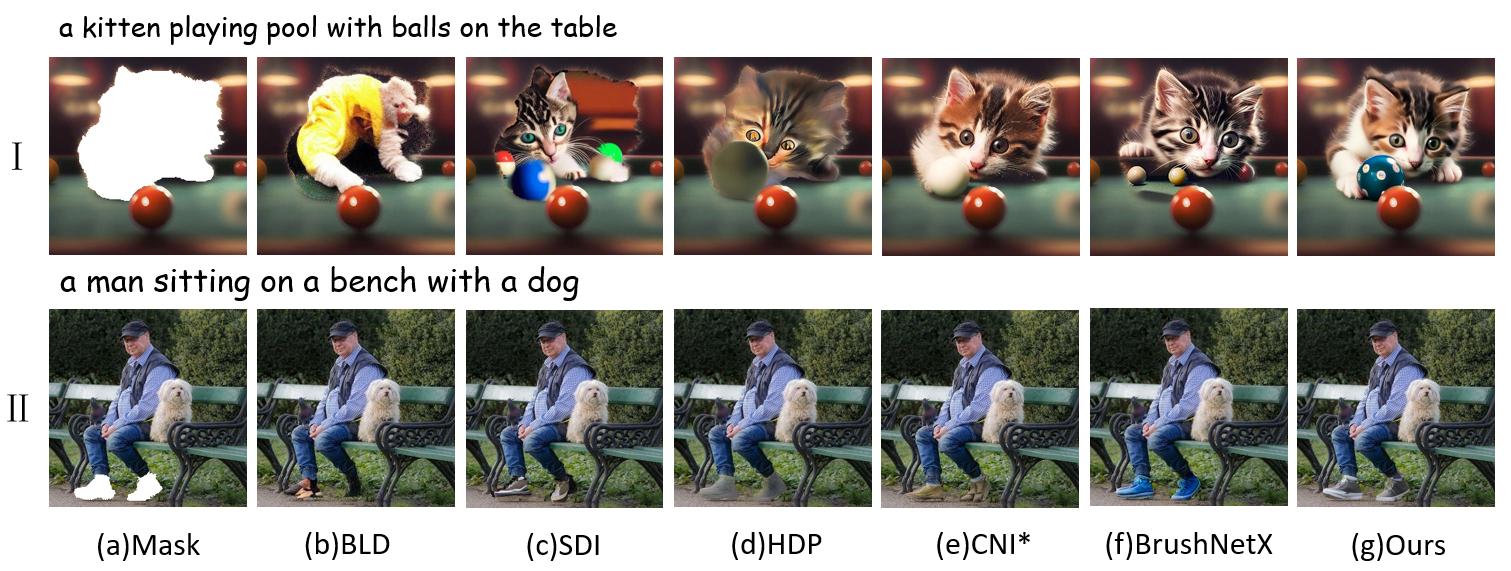}
    \caption{Quanlitative comparisons of the performance of SmartFreeEdit and previous image inpainting methods in nature images. The comparison includes Blended-Diffusion (BLD), Stable Diffusion Inpainting (SDI), HD-Painter (HDP), ControlNet Inpainting (CNI), and the optimized BrushNetX in BrushEdit. }
    \label{fig:inpainting}
    \Description{}
\end{figure*}

\begin{table*}[htbp]
  \caption{Ablation study on the effectiveness of reasoning segmentation and HyperGraph in SmartFreeEdit.}
  \label{tab:ablation}
  \begin{tabular}{c|ccccc|ccccc}%l
    \toprule
    % \multirow{2}{*}{Methods} & \multicolumn{5}{c}{Understanding Scenarious}  & \multicolumn{5}{c}{Reasoning Scenarios}\\
\multirow{2}{*}{Methods}&
\multicolumn{5}{c}{Understanding Scenarios} & \multicolumn{5}{c}{Reasoning Scenarios} \\
%\cmidrule(lr){2--11}
    \cmidrule(lr){2-11} 
  &PSNR↑&\(LPIPS_{\times 10^{3}}\downarrow\)&SSIM↑&CLIPSim↑&Ins-align↑&PSNR↑&\(LPIPS_{\times 10^{3}}\downarrow\)&SSIM↑&CLIPSim↑&Ins-align↑\\    
    \midrule
Baseline&27.92	&68.25&0.72&21.91	&0.51&
    \textbf{\color{red}{32.75}}& \textbf{\color{red}{46.00}}&	0.89&20.21&	0.51\\

  +ReSeg&\textbf{\color{red}{30.34}}	&54.92&0.91&23.76	&0.86&
    32.41& 	65.00&	0.90&21.02&	0.70\\

  +HyPConv&28.99&\textbf{\color{red}{51.49}}&\textbf{\color{red}{0.92}}&\textbf{\color{red}{24.43}}&\textbf{\color{red}{0.86}}&31.27&47.00&\textbf{\color{red}{0.92}}&\textbf{\color{red}{21.46}}&\textbf{\color{red}{0.70}}\\
  \bottomrule
\end{tabular}
\end{table*}

Table \ref{tab:BrushBench} shows quantitative results on the BrushBench benchmark. Our model excelled in image quality, occlusion area retention, and text alignment, particularly scoring highest on IR, HPS, and AS, indicating that the resulting images are not only realistic, but also more attractive to human observers. In terms of regional fidelity, we also performed well on LPIPS and MSE, indicating a high level of consistency between the image and the original.

Although PSNR is slightly lower than BrushNetX, this may be due to PSNR's focus on pixel-level similarity, while BrushNetX optimizes details and improves pixel consistency. Nevertheless, our model is superior on perception-related metrics and has a higher Clip value, confirming that the addition of hypergraph computation not only improves the structure and visual quality of the image, but also significantly enhances the model's ability to understand and complete text instructions in a fine-grained and controlled manner.

Figure \ref{fig:inpainting} shows how SmartFreeEdit qualitatively compares to other leading methods in a real-world scenario. In this figure, our SmartFreeEdit produces a natural and detailed cat that blends seamlessly with the unmasked area and successfully understands and fixes the "shoe" instruction and does a global fix, while BrushNetX only focuses on the surrounding parts of the shoe. Our model has significant advantages in following text instructions and detail recovery, and can accurately process complex image content; In contrast, other methods often produce blurry or incorrect details that fail to adequately restore realism and consistency to the image.

\balance

\subsection{Ablation Study}

To verify the validity of reasoning segmentation (ReSeg) and Hypergraph Convolution (HyPConv) in the SmartFreeEdit framework, we conducted an ablation study under the understanding and inference editing scenarios. As shown in Table \ref{tab:ablation}, we compared the performance of three variants of the system:
1) Baseline: A mask is generated using Grounded-SAM \cite{ren2024grounded} and processed using BrushNet image editing model, representing the existing MLLM bootstrapped image editing pipeline. 2) We replace Grounded-SAM with ReSeg to generate more accurate and semantically consistent mask of instruction. 3) In the full version, the HyPConv is added to enhance the structural consistency and fine-grained realism of the editing area.

In all of the evaluation measures, we observed an improvement in consistency when combined with the proposed modules. Especially in the two scene types, ReSeg significantly improves semantic alignment and editing success, while the addition of HyPConv further improves structural consistency and semantic alignment. Although more modification edits resulted in a slight reduction in PSNR, the image quality was not affected.

% \section{ Limitations and Future Works

\section{Conclusion}
We present SmartFreeEdit, a novel framework that harnesses MLLMs to comprehensively interpret user instructions while preserving spatial relationships and contextual semantics in images by eliminating the need for iterative mask computations and employing HyperGraph-based reasoning to enhance regionally edited content. 
% We present SmartFreeEdit, a novel framework that harnesse MLLMs to comprehensively interpret user instructions while preserving spatial relationships and contextual semantics in images. Our hierarchical architecture employs HyperGraph-based reasoning to enhance regionally edited content, ensuring precise alignment with user intent and strict adherence to real-world physical constraints. By eliminating the need for iterative mask computations, the approach achieves state-of-the-art performance on both BrushBench and Reason-Edit benchmarks. The framework demonstrates unprecedented capability in balancing pixel-level editing accuracy with holistic image fidelity, establishing new standards for intelligent image manipulation without compromising structural consistency.

% then add the following to balance the last page (2 even length columns).
% If you have used \usepackage{balance} include \balance between \bibliographystyle & \bibliography
%\bibliographystyle
%\balance
%\bibliography


% \begin{acks}
% \end{acks}

\clearpage
\bibliographystyle{ACM-Reference-Format}
\bibliography{reference}

\appendix







\end{document}